\documentclass[10pt, a4paper]{article}
\usepackage{lrec2022} 
\usepackage{multibib}
\newcites{languageresource}{Language Resources}
\usepackage{graphicx}
\usepackage{tabularx}
\usepackage{soul}
\usepackage{titlesec}
\titleformat{\section}{\normalfont\large\bfseries\center}{\thesection.}{1em}{}
\titleformat{\subsection}{\normalfont\SmallTitleFont\bfseries\raggedright}{\thesubsection.}{1em}{}
\titleformat{\subsubsection}{\normalfont\normalsize\bfseries\raggedright}{\thesubsubsection.}{1em}{}
\renewcommand\thesection{\arabic{section}}
\renewcommand\thesubsection{\thesection.\arabic{subsection}}
\renewcommand\thesubsubsection{\thesubsection.\arabic{subsubsection}}

\usepackage{epstopdf}
\usepackage[utf8]{inputenc}

\usepackage{hyperref}
\usepackage{xstring}

\usepackage{color}

\title{Simplifying Semantic Annotations of SMCalFlow}

\name{Joram Meron} 

\address{Telepathy Labs GmbH \\
         36 Militärstrasse, Zurich, Switzerland  \\
         joram.meron@telepathy.ai\\}

\abstract{
SMCalFlow \cite{andreas2020task-oriented} is a large corpus of semantically detailed annotations of task-oriented natural dialogues. The annotations use a dataflow approach, in which the annotations are programs which represent user requests. Despite the  availability, size and richness of this annotated corpus, it has seen only very limited use in dialogue systems research work, at least in part due to the difficulty in understanding and using the annotations. 
To address these difficulties, this paper suggests a simplification of the SMCalFlow annotations, as well as releases code needed to inspect the execution of the annotated  dataflow programs, which should allow researchers of dialogue systems an easy entry point to experiment with various dataflow based implementations and annotations. }

\begin{document}

\maketitleabstract

\section{Introduction}

As in many other natural language processing tasks, dialogue systems have achieved impressive advances due to the use of machine learning techniques. These techniques typically require large amounts of high quality annotated data in order to ensure that the resulting models will be able to generalize correctly to unseen input. 

Since the models used by dialogue systems need to also learn the effect of previous turns in the dialogue context (as opposed to models which operate on isolated sentences), even larger amounts of training data are needed.

Training data for dialogue systems typically includes the natural language utterances of the user ("request") and agent ("answer"), as well as some structured data representing the state of the dialogue after the turn (including any additional actions affected by the agent). While the user input can be collected from naive users (e.g. using crowd sourcing platforms), the agent response (both natural language and structured data) need skilled annotators which have been trained specifically for the task.

Due to these difficulties, the number , and size, of available datasets for training dialogue systems has been very limited - a few hundreds or thousands of dialogues only, limiting the type of models which can be used. MultiWOZ \cite{budzianowski-etal-2018-multiwoz},  with 10K dialogues and 70K turns was until recently the largest available set, and is being widely  used in many research works.

More recently, SMCalFlow \cite{andreas2020task-oriented} was released, comprising of more than 40K dialogues (totalling more than 155K turns) of natural (non-scripted) task-oriented user-agent interactions in several domains (calendar events, weather, places and people), with semantically rich annotation. 

The Dialogues were collected via a Wizard-of-Oz process.  At each turn, a crowdworker acting as the user was presented with a dialogue as context and asked to append a new utterance. An annotator acting as the agent labelled the utterance, and then selected a natural-language response from a set of candidates produced by the language generation model. Annotators were provided with detailed guidelines containing example annotations and information about available library functions. 

Despite the size of this dataset, and the high level of detail given by the annotations, it was not adopted by the dialogue systems research community. The assumption in this paper is that this is the result, at least in part, of the difficulty in understanding and using this dataset by the research community. This difficulty is due to two factors: 1) The annotation scheme is complex, and lacks sufficient documentation to explain it, and 2) tools to inspect and verify that the annotations are correct were not released.

This paper\footnote{Published in the Proceedings of the 18th Joint ACL - ISO Workshop on Interoperable Semantic Annotation within LREC2022} addresses these difficulties by 1) suggesting a simplified annotation scheme, which, hopefully is easier to understand, and 2) releasing the necessary code to inspect the annotation results. It is hoped that with these contributions, the research community will be encourage to explore and exploit the potential of this rich dataset.

\section{Dataflow Dialogues}

SMCalFlow uses dataflow (DF) computational graphs, composed of a rich set of both general and application specific functions (see figures \ref{ex1-org-txt} and \ref{ex2-org-txt}),  to represent the user requests as rich compositional (hierarchical) expressions. These computational graphs can be executed, which results in manipulating the computational graphs, generating an answer (possibly an error message), and optionally producing some side effects through API's to external services (e.g. updating the user's calendar appointments on an external database).

The prominent features of this paradigm are:
\begin{itemize}
\item The dialogue history is represented as a set of graphs, where each computational graph typically represents one user turn.
\item It has a {\it refer} operation to search over the current and previous computational graphs (as well as external resources) which allows easy look-up and re-use of graph nodes which occurred previously in the dialogue.
\item It has a {\it revise} operation which allows modification and reuse of previous computations
\item It has an exception mechanism which allows convenient interaction with the user (e.g. asking for missing information, and resuming the computation once the information is supplied).
\end{itemize}

These features correspond to essential phenomena in natural conversations (referring to previous turns, modifying previous requests, reacting to wrong information, etc.), which allows the system to effectively handle these kinds of user requests.

\section{Simplifying SMCalFlow}

In this work, a simplified annotation is presented, with the motivation to reduce the effort on the annotator/reader, without increasing the learning effort for the machine translation models used to convert the users' natural language requests to the target annotation format. 

As described below, this simplification requires some additional logic  to be implemented in the execution engine, as well as in the individual functions, but this additional logic is typically trivial. 

The starting point of this work is SMCalFlow, with its original annotation style.  Because of its size, and the limited resources available in this work, manual modification of individual annotations were not feasible. Instead, the modifications had to be done fully automatically, using a programmatic solution to do the conversion. The consequences of this decision are:
\begin{itemize}
\item The new annotations are still tied to the original ones, so some of the design decisions made by the original annotators are difficult to change (as opposed to the case where the new annotation would start from scratch). 
\item Specifically, any mistakes or anomalies in the original annotations are carried over to the simplified annotations.
\item An automatic conversion mechanism had to be created and configured to convert the annotations correctly.
\end{itemize}

While DF is not inherently complicated, finding a good design is a challenging task. A novel aspect of this challenge is the need for the design to function correctly within the DF paradigm (e.g. use the {\it refer} and {\it revise} operators). Indeed, one of the motivations of this work is the hope that the community can suggest interesting new designs, which can serve as templates for further applications.

\subsection{Simplification Mechanism}

The simplification was performed by implementing a set of tree transformation rules, which convert specified sub-trees of the original expressions into simplified sub-trees. The transformation code is part of the release, and can be used to replicate the work reported here.

The simplification is applied to the whole dataset, resulting in a simplified dataset, which can then be fed into the exact same machine translation training and evaluation pipeline used in the original paper.

For convenience, the simplified format uses Python style expressions (as opposed to the Lisp style S-expressions in the original dataset), as this format is generally more familiar (the released system itself is written in Python).

\subsection{\label{simp-principles}Simplification Approach}

The design principles for the simplifications were: 
\begin{enumerate}
\item Retain only necessary information
\item Avoid explicit logical steps
\item Move logic from the annotation to the implementation of the individual functions
\item Group and reuse repeating sequences of functions
\item Relax strict type constraints
\item Reduce unnecessary compositions
\end{enumerate} 

Practically this means: Try to omit any information which can be {\it deterministically} inferred - keep only information which can not be inferred. Specifically, logical steps which can be inferred from context, are moved from the annotation into the implementation of the functions. For example:

\begin{itemize}
\item Explicit type casts which are clear from the context can be omitted.
\item When needed information is missing in the user input, but can be inferred from the computation, the simplified annotation should leave the inference of the missing information to the function implementation.
\item The simplified annotation tries to avoid fragments of the original annotation which serve only "formal" purposes, and instead tries to style the annotation to be closer to a more natural/comprehensible description of the user requests (and in general be closer to the surface form of the user request, as can be seen in the examples). 
\end{itemize}

Below are examples of original vs. simplified annotations.

\subsubsection{Example 1}

The user's request is: 

{\tt "Delete the meeting with John's supervisor tomorrow"}.

\begin{figure}
\includegraphics[height=0.17\textheight,trim= 0.7in 0.4in 0.5in 0.1in]{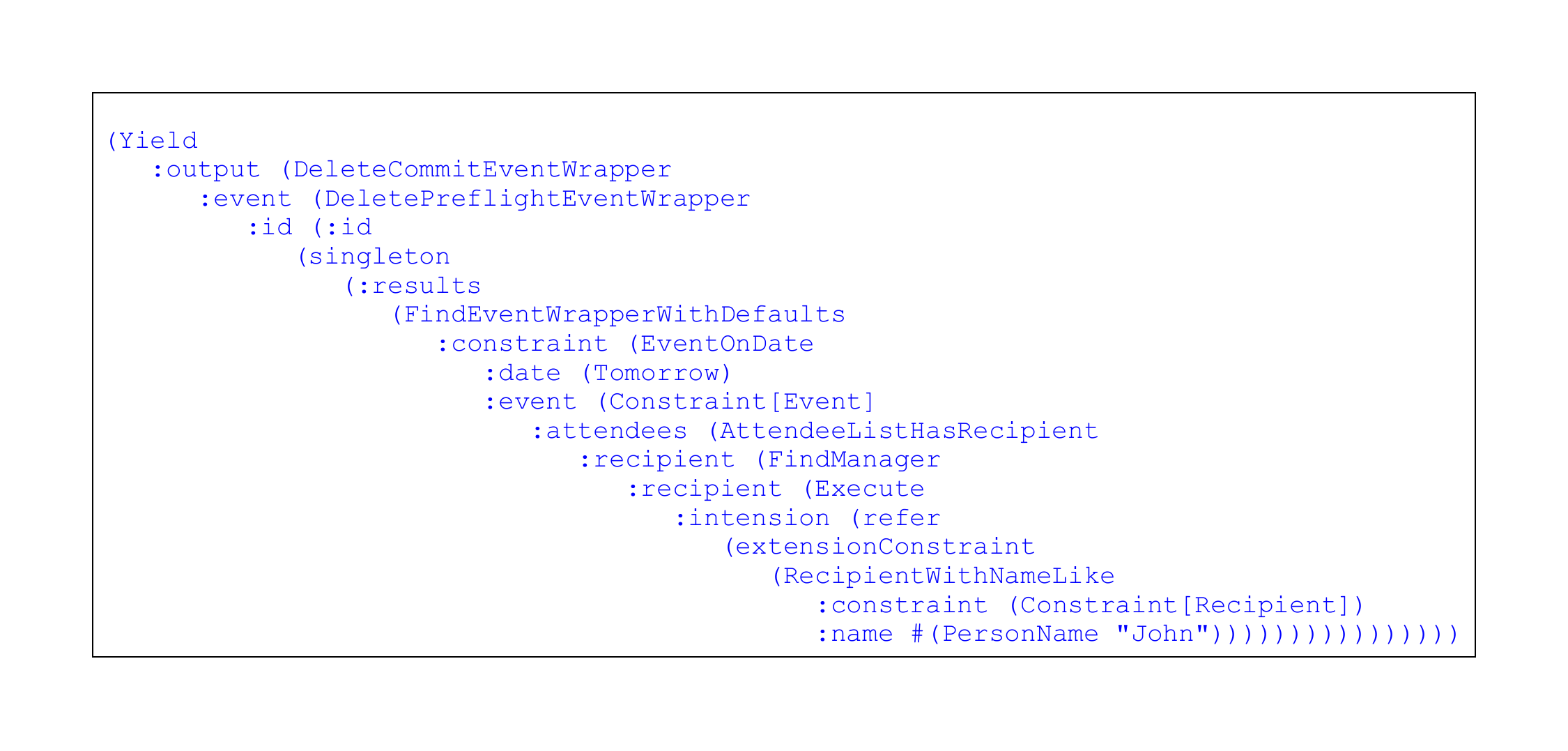}
\caption{\label{ex1-org-txt} Example 1  - original annotation}
\end{figure}

\begin{figure}
\includegraphics[height=0.1\textheight,trim= 0.1in 0.4in 0.0in 0.1in]{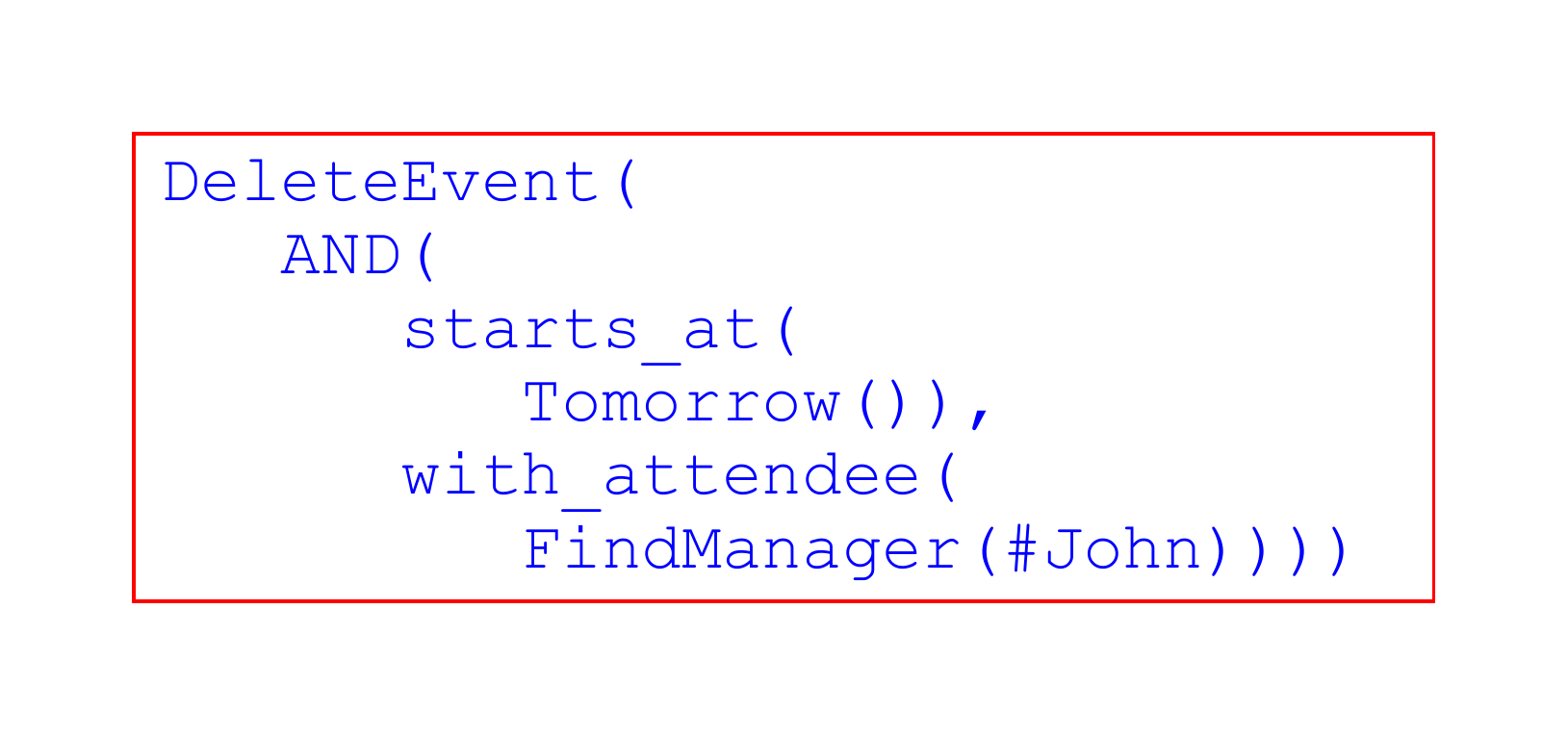}
\caption{\label{ex1-simp-txt} Example 1 - simplified annotation}
\end{figure}

\begin{figure}
\includegraphics[height=0.5\textheight,trim= 0.3in 0.6in 0.43in 0.58in]{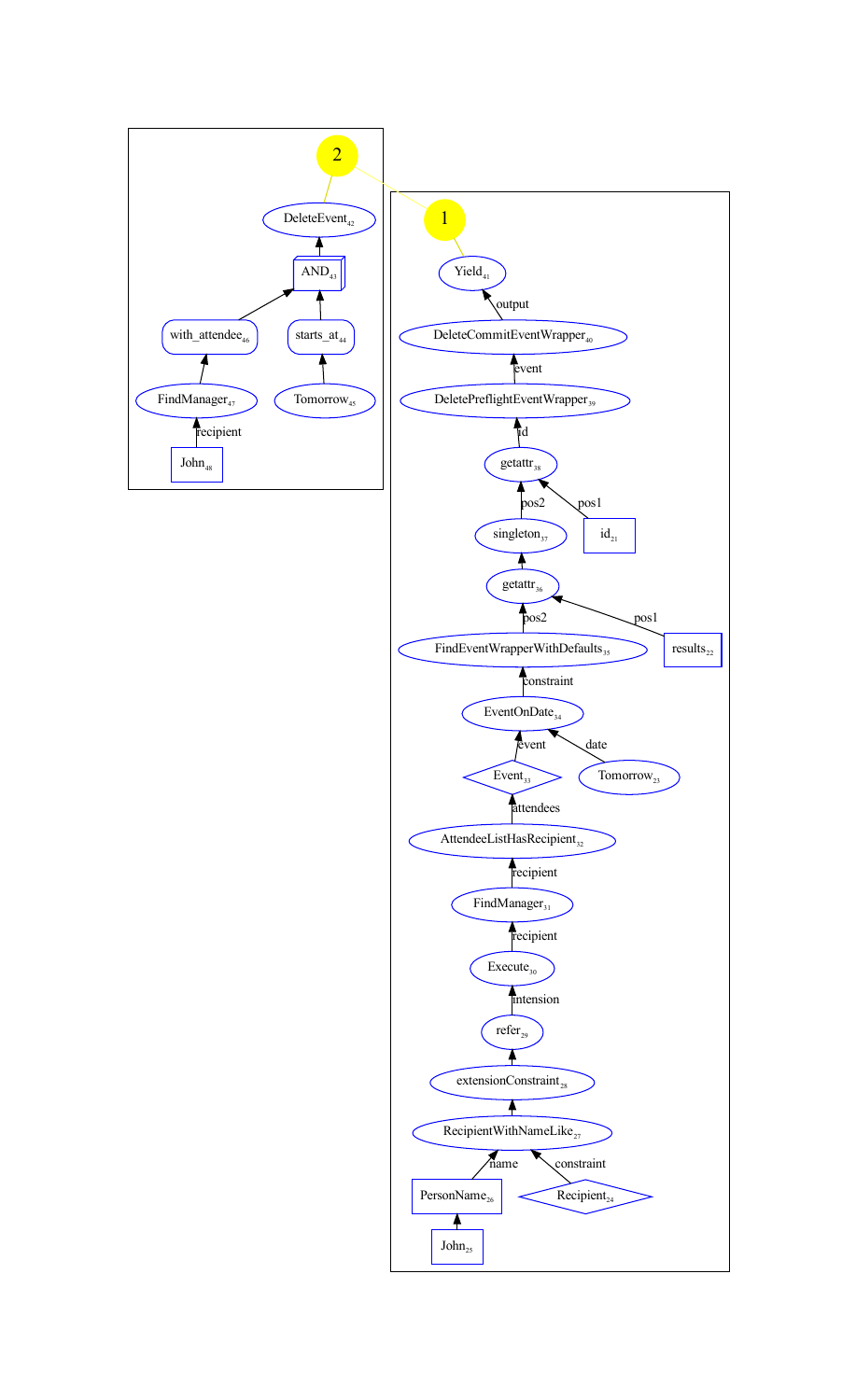}
\caption{\label{ex1-grp-txt} Example 1 - original and simplified annotations shown as graphs}
\end{figure}

Figures \ref{ex1-org-txt} and \ref{ex1-simp-txt} show the original and simplified annotations for this request.  Figure \ref{ex1-grp-txt} show the annotations as computational graphs.

This example illustrates a few of the simplification ideas:
\begin{itemize}
\item Computational steps which always appear together are bundled into one step: {\it 'DeletePreflightEventWrapper'} and {\it 'DeleteCommitEventWrapper'} correspond to two sub steps of the act of deleting an event. Here, they are simplify by combining them into one step {\it 'DeleteEvent'}.
\item Relax strict type constraints in the annotation. In the original annotation, {\it 'DeletePreflightEventWrapper'} can accept only an integer input (representing the unique id of the event to be deleted). In the simplified version, the implementation of the {\it 'DeletePreflightEventWrapper'} can handle additional types of input, by calling the necessary type conversion, i.e.: if the input is an {\it 'Event'} type, then extract its {\it 'Event.id'} value, and if the input is a set of events, then additionally invoke a call to the {\it 'singleton'} function.
\item Avoid explicit logical steps. In the original annotation, the process of searching for a person is an explicit part of the annotation (see the input to the {\it 'FindManager'} function). In the simplified annotation, this logic is added to the implementation of {\it 'FindManager'} , so the annotation can be simply {\it 'FindManager(John)}.
\item Avoid unnecessary compositions and annotations which serve only "formal" purpose. In the original annotation,  {\it 'RecipientWithNameLike'}  implements a compositional pattern, where one of the inputs is an empty constraint, which is dropped in the simplified annotation (in this case, the whole surrounding block is also removed).
\end{itemize}

\subsubsection{Example 2}

The user's request is:

 { \tt "I want John, Emily, John's supervisor and Bob to attend"}. 

\begin{figure}
\includegraphics[height=0.35\textheight,trim= 0.8in 0.2in 0.1in 0.2in]{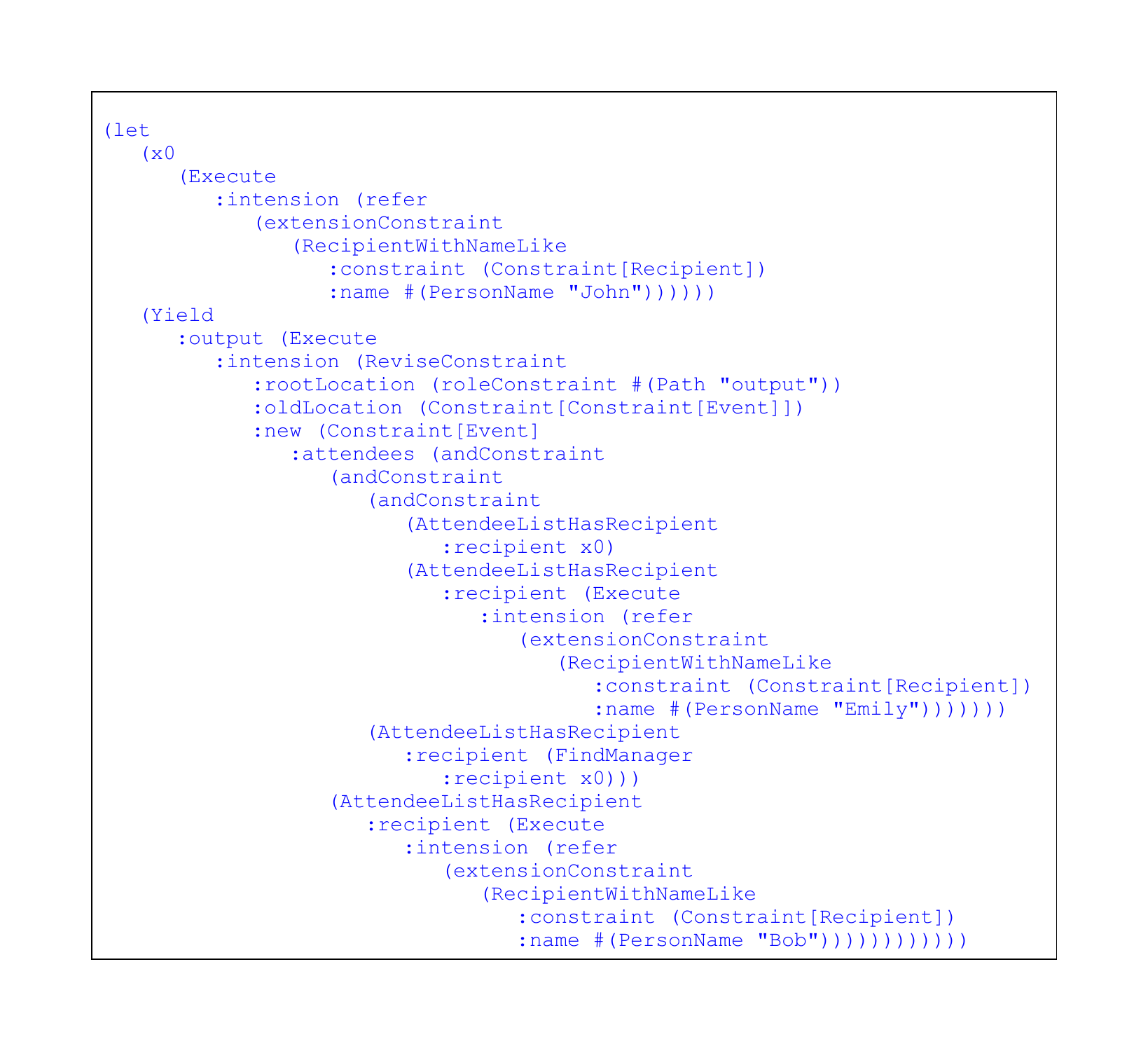}
\caption{\label{ex2-org-txt} Example 2  - original annotation}
\end{figure}

\begin{figure}
\includegraphics[height=0.17\textheight,trim= 0in 0in 0in 0in]{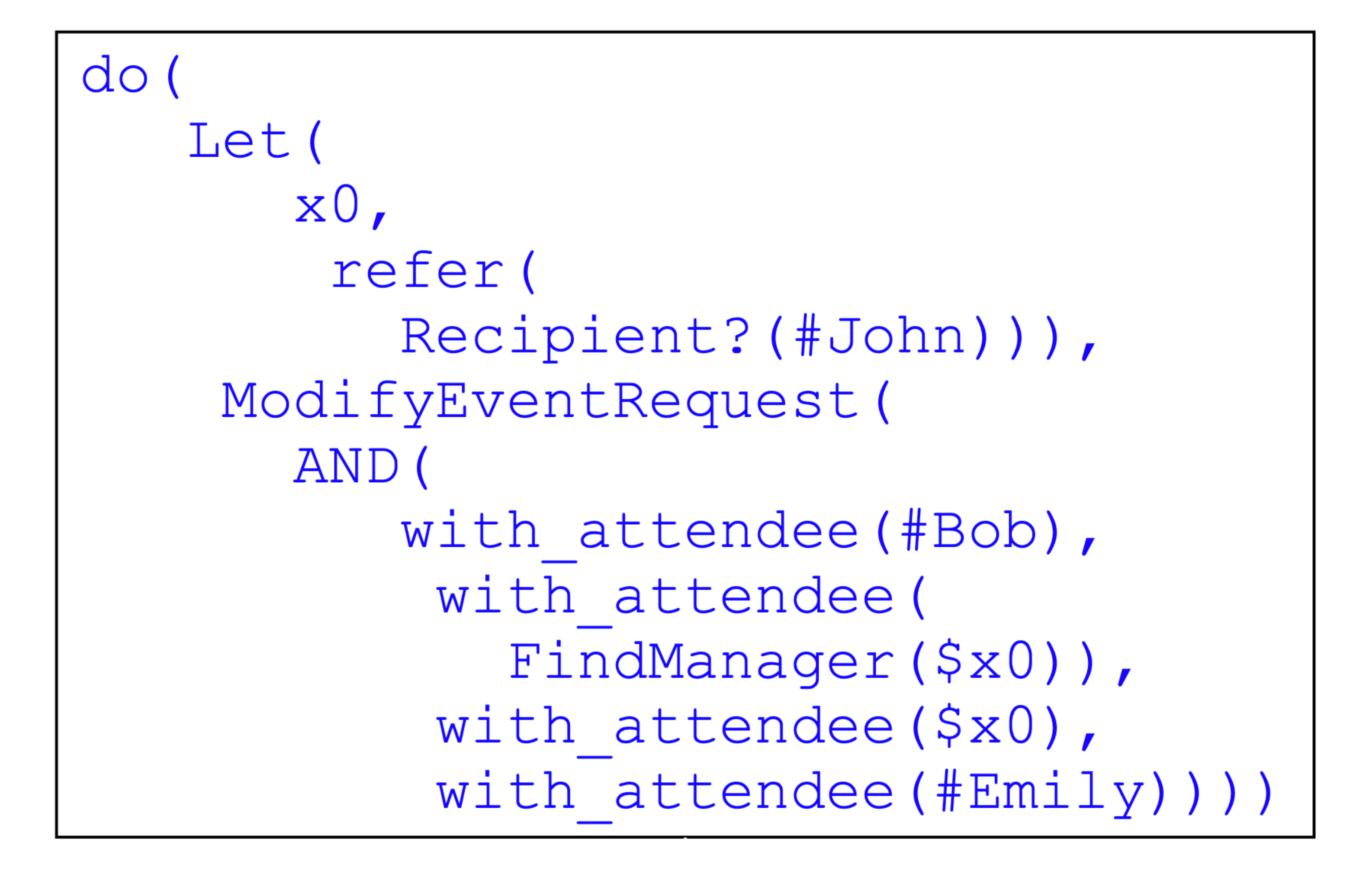}
\caption{\label{ex2-simp-txt} Example 2  - simplified annotation}
\end{figure}

Figures \ref{ex2-org-txt} and \ref{ex2-simp-txt} show the original and simplified annotations for this request.  

This example illustrates some simplification ideas:
\begin{itemize}
\item Simplification of the assignment construct, avoiding unnecessary assignments (which are used only once) - in this example, variable {\it x1} is used only once, so it is substituted directly into the main expression.
\item Reducing the use of compositions, in favour of flatter expressions. In this example,  instead of chaining constraints using {\it 'andConstraint'}, the simplified annotation uses a flat {\it 'AND'} construct.
\end{itemize}

\begin{figure*}
\includegraphics[height=0.45\textheight,trim= 0.0in 0.7in 0.2in 0.7in]{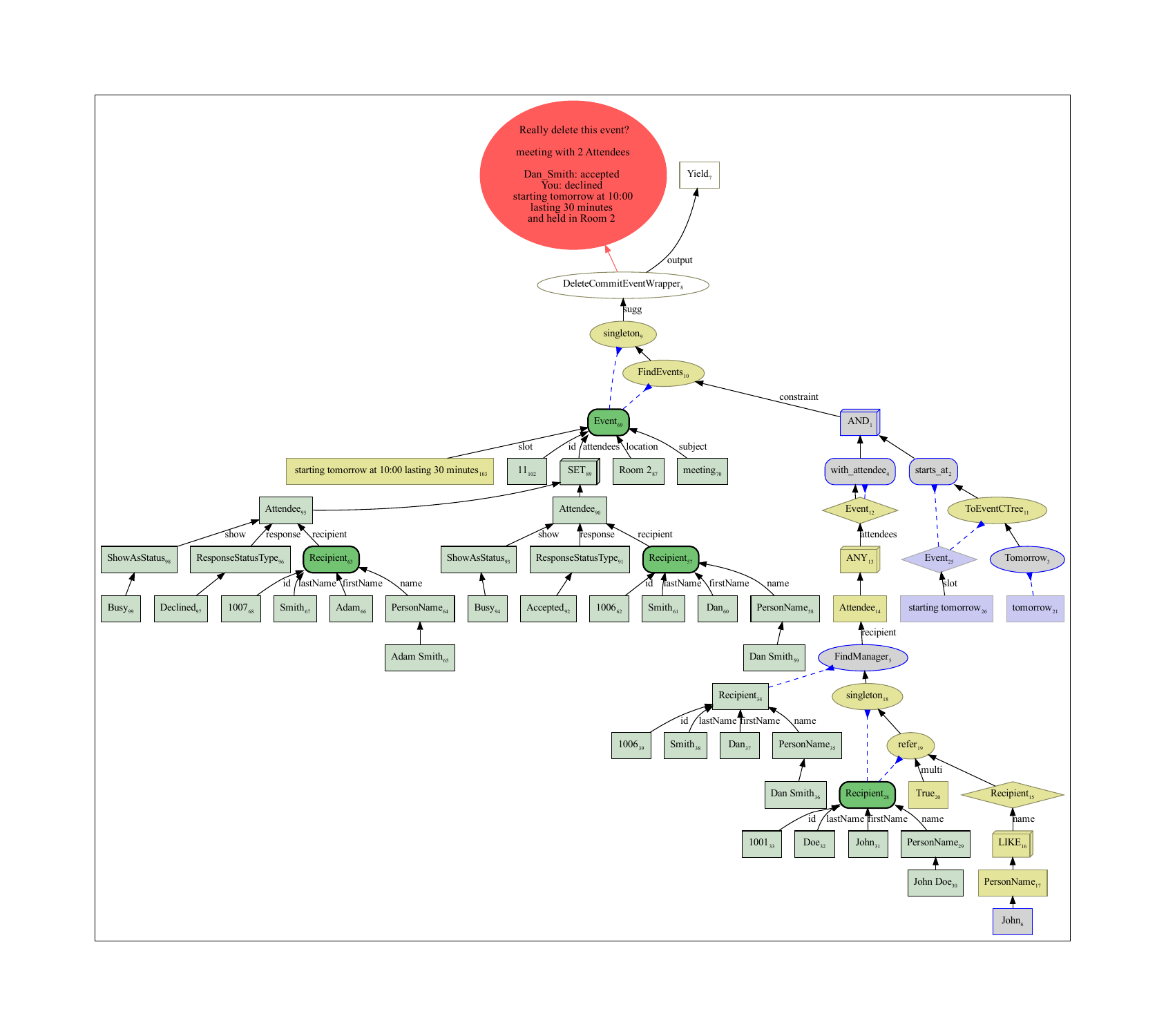}
\caption{\label{ex1-exec} The result of transforming and evaluating the simplified annotation for the request {\it "Delete the meeting with John's supervisor tomorrow"}. Nodes from the simplified annotation are shown in gray. Yellow nodes indicate nodes added automatically by expansion logic, green nodes indicate information extracted from external DB (the result of searching for John and his manager, and for the event matching the requested constraints). Blue dashed  lines indicate execution results (some nodes are omitted to reduce clutter).}
\end{figure*}

\subsection{Executing Simplified Annotations}

At execution time, an additional step transforms the simplified annotation to a fully executable expression. This is done, again, by implementing tree transformation rules (for each function), which can add deterministically inferable missing information/steps (e.g. casting input to the right type, or performing other conversions/functions based on input type). 

This step could be viewed, in principle, as the inverse of the dataset simplification step, but in practice the run-time transformation of the simplified annotation is often  quite different  from the original annotation, due to different design decisions and function implementations. 

Figure \ref{ex1-exec} shows the result of transformation and execution of the simplified annotation for example dialogue 1 above. The transformed graph is clearly different from the original annotation's graph.

\subsection{Simplification Results}

Since the original code to execute SMCalFlow was not released (and documentation not supplied), it is impossible to verify that the suggested simplifications implement/execute the exact same logic (in fact this was one of the motivations for this paper). It can only be left to the readers to inspect the simplified annotations and the code and draw their own conclusions. 

Qualitative evaluation confirmed correct execution of a sample of expressions, but further work is needed to obtain more significant quantitative evaluation.

\begin{table}
\begin{center}
\begin{tabularx}{\columnwidth}{|l|X|}

\hline
  & Program Length \\
\hline
Original Annotation & (11, 37, 58) \\
\hline
Simplified Annotation & (2, 11, 20) \\
\hline
\end{tabularx}
\caption{\label{program-length}
Program length of the two annotation styles. Length is measured as number of seq2seq target tokens, when translating user request to annotation. Showing (.25, .50, .75) quantiles over the entire dataset.}
\end{center}
\end{table}

Table \ref{program-length} shows the results of a comparison of the annotation lengths of the original and simplified annotations, confirming that the simplification does make the annotation significantly shorter. 

The example annotations shown above should show that the simplified annotations are not just significantly shorter, but are also significantly simpler to understand, which should reduce annotation efforts when creating new training data.

Table \ref{sub_data} shows that the simplification did not reduce (and maybe increased) the accuracy of machine translation of natural language user requests to dataflow expressions. Refinement of the simplification rules may result in further improvements.

\begin{table}
\begin{center}
\begin{tabularx}{\columnwidth}{|c|c|c|c|c|}
\hline
  & {\small 1k} & {\small 3k} & {\small 10k} & {\small 33k}\\
\hline
{\small Original} & 30.2{\small \textpm 3.6} &  41.8{\small \textpm 7.9} & 55.7{\small \textpm7.0} & 72.8 \\
\hline
{\small Simplified} & 35.9{\small \textpm 4.2} & 47.7{\small \textpm 7.0} &  62.1{\small \textpm 1.9} & 73.8 \\
\hline
\end{tabularx}
\caption{\label{sub_data}
Translation accuracy (exact match) as function of training data size, showing average and std. (in percent) over 7 randomly selected samples per size. }
\end{center}
\end{table}

\section{Further Work}

The work presented in this paper is still in progress, trying to improve the simplified annotation format and the automatic simplification.

Accordingly, the implementation of the executable functions will continue to evolve, to be able to correctly execute modified annotation formats.

While the automatic simplification covers all of the dataset, the implementation of functions has concentrated mostly (but not exclusively) on turns dealing with the calendar domain (which is the most complex domain in this dataset).

Further ideas and work on the simplified annotation definition (and transformation process) from the community are encouraged. With the released code, researchers should be able to experiment with new ideas and share them with the community. 

Additional areas of interest may include:
\begin{itemize}
\item 
Evaluation: in addition to the exact-match metric for translation accuracy, other metrics can be used, such as comparison of execution results, graph structure similarity, etc. 

\item 
Using the graph structure: the graph structure (at different points of the execution) can be used by prediction models. 

\item 
Different design patterns which are beneficial to specific parts of the system. For example, the execution of a computation graph could emit various types of information which would then be useful for subsequent prediction models.

\end{itemize}

\section{Conclusion}

A simplification of the  SMCalFlow annotations has been presented. Some simplification principles have been suggested, and an automatic conversion tool has been implemented. Examples have been given to show that the simplified annotations are significantly shorter, as well as easier to understand, than the original annotations.

The code for reproducing this work \footnote{https://github.com/telepathylabsai/OpenDF} allows to run annotation simplification as well as executing these annotations to inspect and verify they satisfy user requests, and should lower the barrier of entry into Dataflow dialogue design for interested researchers, allowing them to experiment with new ideas.

\section{Bibliographical References}\label{reference}

\bibliographystyle{lrec2022-bib}
\bibliography{lrec_df}

\end{document}